\pdfoutput=1

\documentclass[11pt]{article}

\usepackage[final]{acl}

\usepackage{times}
\usepackage{latexsym}
\usepackage{booktabs}
\usepackage{amsmath}
\usepackage{amssymb}
\usepackage{multirow}
\usepackage{pifont}
\usepackage{bbding} 
\usepackage[T1]{fontenc}

\usepackage[utf8]{inputenc}

\usepackage{microtype}

\usepackage{graphicx}

\title{MadaKV: Adaptive Modality-Perception KV Cache Eviction for Efficient Multimodal Long-Context Inference}

\author{
    \bf Kunxi Li\textsuperscript{\rm 1}\thanks{These authors contributed equally.}, 
    Zhonghua Jiang\textsuperscript{\rm 1}\footnotemark[1], 
    Zhouzhou Shen\textsuperscript{\rm 2}, 
    Zhaode Wang\textsuperscript{\rm 3}, 
    Chengfei Lv\textsuperscript{\rm 3}, \\ \bf
    Shengyu Zhang\textsuperscript{\rm 1}\thanks{Corresponding author.}, 
    Fan Wu\textsuperscript{\rm 4},
    Fei Wu\textsuperscript{\rm 1} \\
    \textsuperscript{1}Zhejiang University, \textsuperscript{2}Southeast University, \textsuperscript{3}Alibaba, \textsuperscript{4}Shanghai Jiao Tong University \\
    \footnotesize\texttt{\{kunxili, jiangzhonghua, sy\_zhang, wufei\}@zju.edu.cn}, 
  \texttt{zhouzhoushen@seu.edu.cn}, \\ \footnotesize\texttt{\{zhaode.wzd, chengfei.lcf\}@alibaba-inc.com}, \texttt{fwu@cs.sjtu.edu.cn}
}

\begin{document}
\maketitle
\begin{abstract}
This paper introduces MadaKV, a modality-adaptive key-value (KV) cache eviction strategy designed to enhance the efficiency of multimodal large language models (MLLMs) in long-context inference. In multimodal scenarios, attention heads exhibit varying preferences for different modalities, resulting in significant disparities in modality importance across attention heads. Traditional KV cache eviction methods, which are tailored for unimodal settings, fail to capture modality-specific information, thereby yielding suboptimal performance. MadaKV addresses these challenges through two key components: modality preference adaptation and hierarchical compression compensation. By dynamically sensing modality information within attention heads and adaptively retaining critical tokens, MadaKV achieves substantial reductions in KV cache memory footprint and model inference decoding latency (1.3 to 1.5 times improvement) while maintaining high accuracy across various multimodal long-context tasks. Extensive experiments on representative MLLMs and the MileBench benchmark demonstrate the effectiveness of MadaKV compared to existing KV cache eviction methods.
\end{abstract}

\section{Introduction}

In recent years, leveraging the transformer architecture, autoregressive language models have shown remarkable progress in handling long-context inputs across various tasks \cite{liu2024world,tworkowski2024focused,touvron2023llama,jiang2023mistral}. 
However, the autoregressive decoding mechanism, which necessitates the consideration of every preceding token when generating a new one, incurs quadratic complexity, posing computational challenges. 
The KV cache approach addresses this by caching key and value tensors from past tokens, 
reducing the decoding process to linear time complexity.
However, as the context length increases, the KV cache's memory footprint increases significantly \cite{shi2024keep}. Building on prior research that highlights the sparsity within attention mechanisms, where only a subset of tokens significantly influences outcomes \cite{zhao2019explicit,beltagy2020longformer,choromanski2020rethinking,wang2020linformer}, recent studies have focused on identifying the importance of each token and discarding the KV cache of less significant tokens \cite{zhang2024h2o,cai2024pyramidkv,chen2024nacl}. 

\begin{figure}[t]
  \centering
  \includegraphics[width=7.5cm]{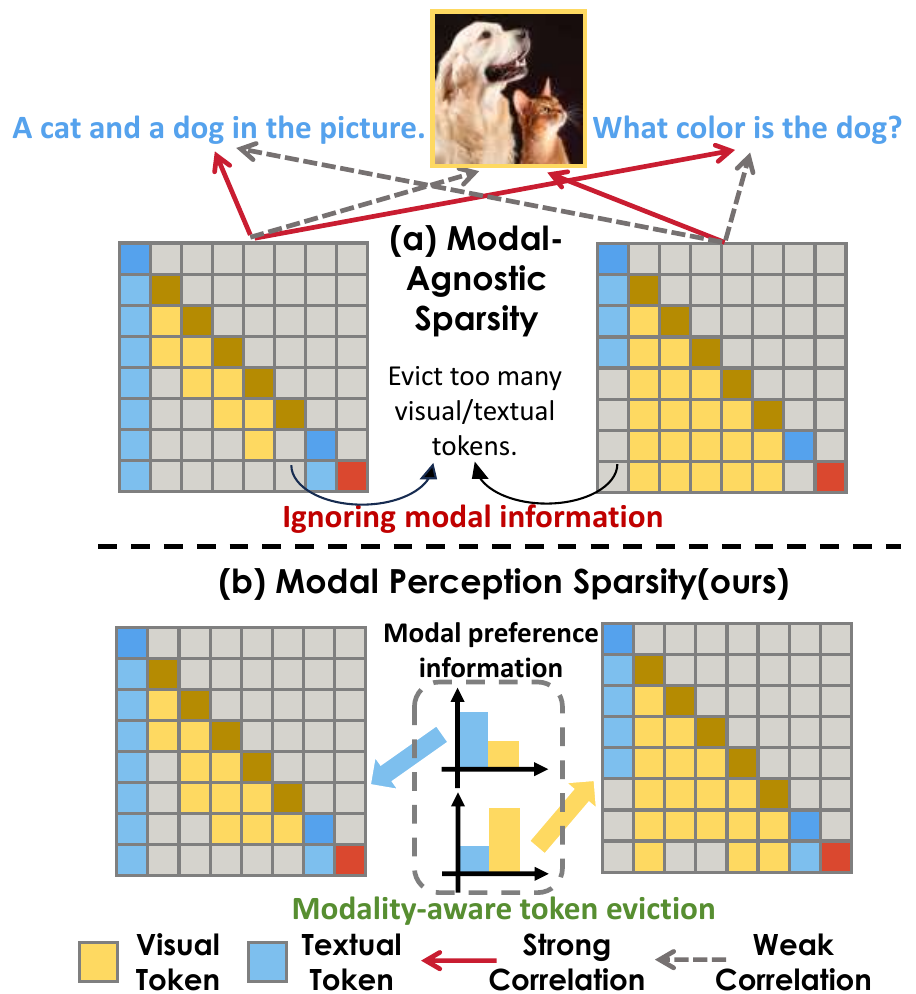}
  \caption{Comparison between Modal-Agnostic Sparsity (a) and MadaKV's 
 Modal Perception Sparsity (b).
}
\label{figure1}
\end{figure}

\begin{figure*}[t]
  \centering
  \includegraphics[width=15.61cm]{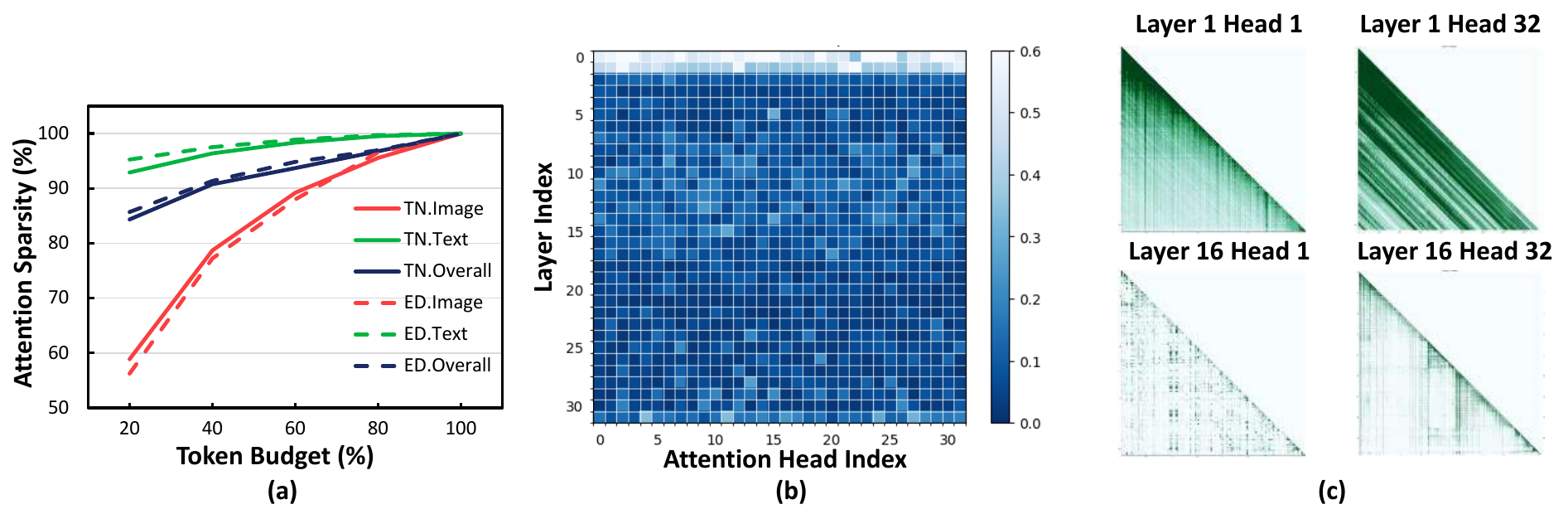}
  \caption{(a) The sparsity differences among tokens of different modalities. TN: Text Needle In A Haystack; ED: Conversational Embodied Dialogue. (b) The distribution of attention scores assigned to text tokens across different attention heads. A higher score indicates that more attention is allocated to text tokens, while a lower score suggests that more attention is directed towards visual tokens. (c) The attention score matrices across different layers.}
  \label{fig:findings}
\end{figure*}

In the field of Multimodal Large Language Models (MLLMs) \cite{liu2024visual,li2024blip,wang2024qwen2}, these challenges are also present, yet traditional single-modality approaches are suboptimal. 
The information density encapsulated within tokens varies significantly across modalities; for instance, textual tokens often encode semantic concepts concisely, while visual tokens require fine-grained spatial representations spanning hundreds of tokens, highlighting the need for modality-specific granularity when evaluating token importance. 
Moreover, the complex interactions between multimodal tokens lead to substantial variations in attention patterns across different input instances. As illustrated in Figure \ref{figure1}, traditional unimodal approaches, lacking awareness of multimodal information, may inadvertently evict critical multimodal tokens, thereby generating erroneous responses.

More recently, LOOK-M \cite{wan2024look} applies the KV cache eviction strategies to MLLMs. However, this approach empirically assigns fixed modality prioritization, overlooking the significance of differences among modalities in various task contexts (examples are provided in the appendix). We analyze the attention scores assigned to visual and text tokens by each attention head on multimodal long-context scenarios. As shown in Figure \ref{fig:findings} (b), the proportion of scores assigned to different modalities varies significantly across attention heads, indicating a preference for modality-specific information. This insight suggests that a KV cache eviction strategy for MLLMs is necessary to accommodate the varying importance of modalities in different contexts.

This paper introduces \textbf{MadaKV}, a plug-and-play modality-adaptive KV cache compression strategy, designed to mitigate the inference costs of MLLMs.     Specifically, our approach comprises two integral components: \textit{modality preference adaptation} (MPA) and \textit{hierarchical compression compensation} (HCC). Firstly, 
MPA dynamically discerns modality preference patterns through real-time analysis of cross-modal contextual interactions, thereby facilitating the learning of specific modality importance for attention head and guiding the eviction of multimodal KV caches accordingly.     Second, recognizing that multimodal attention patterns vary significantly across layers, tasks, and instances, we introduce a progressive hierarchical compression compensation. This mechanism dynamically adjusts the eviction ratio based on the complexity of modality information in the context. Additionally, it incorporates a compensation scheme to balance the cache budget across layers. The compensation is accumulated across layers, summarizing the eviction of modality information from previous layers and providing alignment guidance for the eviction strategy in the current layer. This ensures that the overall cache budget is maintained while preventing error propagation due to modality information compression.

We conduct extensive experiments on representative MLLMs, including LLaVA-v1.5 \cite{liu2024visual} and Qwen2.5-VL, and evaluate their performance across a variety of multimodal long-context tasks within the MileBench \cite{song2024milebench}: temporal multi-image tasks, semantic multi-image tasks, needle in a haystack task, and image
retrieval tasks. MadaKV achieves better accuracy for a given degree of KV cache sparsity than baselines    Specifically, MadaKV achieves a 1.3 to 1.5 times improvement in model inference decoding latency and reduces the KV Cache Memory Footprint by 80\% to 95\%,
while maintaining performance on multimodal long-context tasks.

\section{Related work}

Memory efficient inference has always been an important research direction in the field of deep learning, with early studies mainly focusing on techniques such as activation checkpoints~\cite{chen2016training,zhou2025colacollaborativelowrankadaptation}, offloading~\cite{ren2021zero}, and dynamic memory management~\cite{rhu2016vdnn}. 
Although these methods alleviate memory pressure to some extent, they often introduce additional latency or hardware dependency, limiting their widespread adoption in practical applications. 
For large language models (LLMs)~\cite{floridi2020gpt,achiam2023gpt,meta2024introducing,yang2024qwen2,Kong_Wang_Shen_Zhu_Xu_Su_2025,zhou2025cuffkttacklinglearnersrealtime,zhou2025revisiting} based on transformers~\cite{waswani2017attention,10.1145/3690624.3709275,zhou2025disentangled,zhou2024cuffkt}, KV cache, as a key component in autoregressive generation tasks, stores the key and value vectors calculated in the attention mechanism, thereby avoiding redundant calculations during the decoding process. 
However, as the model size and sequence length increase, the memory usage of KV cache gradually becomes a performance bottleneck~\cite{shi2024keep}. 
Recent research has explored various methods for optimizing KV cache, such as quantization methods~\cite{hooper2024kvquant,yang2024no,sheng2023flexgen} that reduce memory usage by lowering the accuracy of cache values (such as from FP16 to INT8), and eviction strategies~\cite{zhang2024h2o, yang2024pyramidinfer} that selectively delete less important key-value pairs~\cite{xiao2023efficient, han2024lm}. 
Although these techniques have achieved significant results in single-modal models, they are not directly applicable to multimodal models due to the unique characteristics of cross-modal attention.

MLLMs, such as Flamingo~\cite{alayrac2022flamingo}, Qwen-VL~\cite{bai2023qwen}, and LLaVA~\cite{liu2023visual} integrate information from multiple modalities achieving state-of-the-art performance in tasks such as visual question answering and image captioning. 

Due to the fact that MLLMs typically require a large number of tokens to store multimodal inputs~\cite{yin2023survey}, such as images, this introduces additional complexity when managing KV Cache. 
Recent studies have proposed methods such as sparse attention~\cite{child2019generating} and low rank approximation~\cite{song2024low} to reduce the computational overhead of MLLMs, but have not solved the memory usage problem of KV Cache in multimodal settings.

Compressing KV Cache in MLLMs presents unique challenges. Cross-modal attention patterns are often different from traditional LLMs, which limits the effectiveness of conventional KV cache optimization techniques in MLLMs. Some recent studies such as FastGen~\cite{ge2023model} and H2O~\cite{zhang2024h2o} propose optimizing the efficiency of KV Cache usage through lightweight model analysis and adaptive cache compression strategies. LOOK-M~\cite{wan2024look} prioritizes evicting or merging the KV Cache of image tokens by observing cross-modal attention patterns, while significantly compressing the KV Cache while maintaining contextual quality without the need for fine-tuning the model. However, the method of prioritizing the eviction of visual tokens exhibits limited applicability in multimodal settings. Departing from prior work, our method introduces context-aware modality prioritization.  This adaptive mechanism mitigates critical information loss and demonstrates robust effectiveness across a wider spectrum of multimodal scenarios.

\section{Method}

\begin{figure*}[t]
  \centering
  \includegraphics[width=15.61cm]{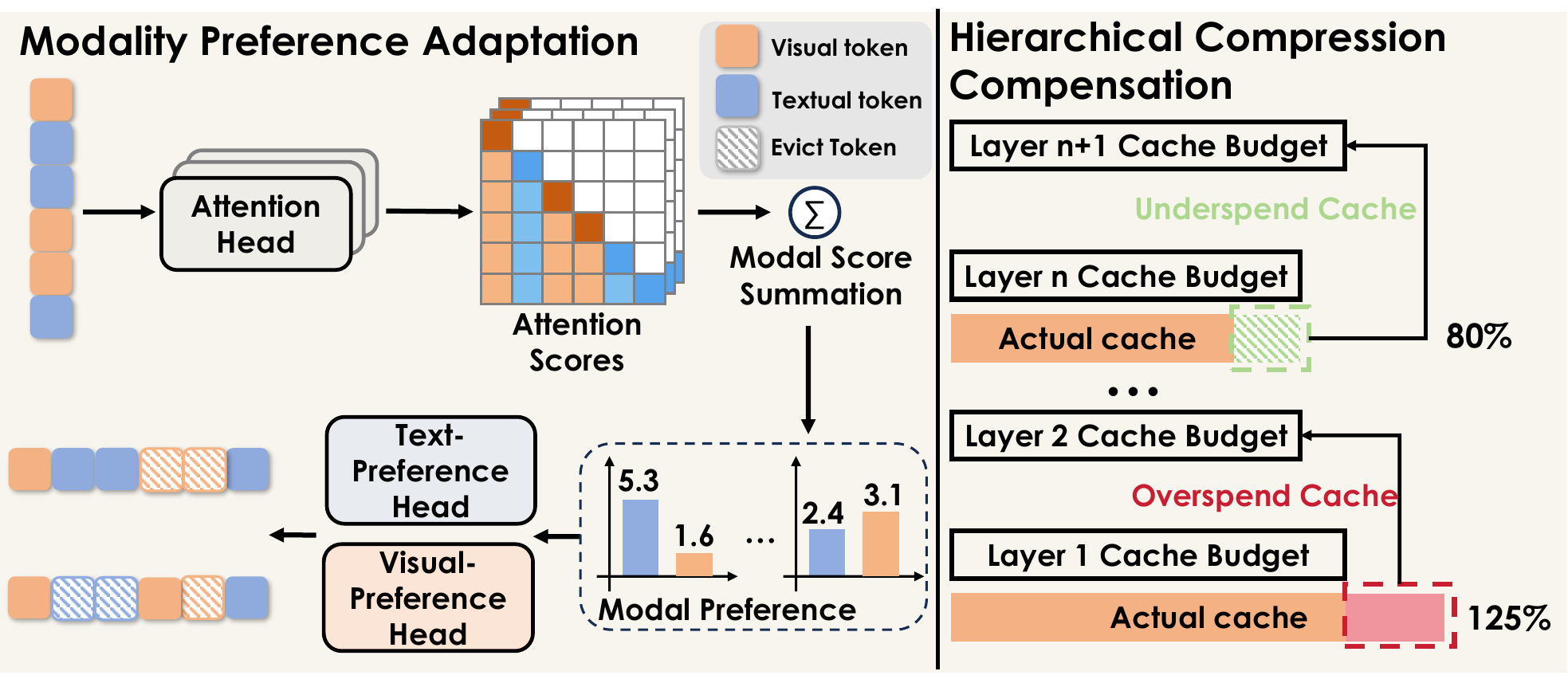}
  \caption{Overview of MadaKV. Modality Preference Adaptation identifies the modality preferences of attention heads and guides the computation of cache budgets for each modality. Hierarchical Compression Compensation employs a compensation mechanism to coordinate budgets across layers.  This component ensures that the trade-offs between compression and information retention are balanced across different layers of the model.}
  \vspace{-0.5cm}
\end{figure*}

\subsection{Preliminary}

In the inference process of MLLMs, KV Cache is a key memory mechanism used to store the Key and Value in the attention mechanism, thereby accelerating the generation process and reducing computational overhead. The use of KV Cache mainly involves two stages: prefilling and decoding.

\noindent \textbf{Prefill phase:} During this stage, the model ingests an input sequence $X=[x_1^t, x_2^t,..., x_{n-1} ^v, x_n ^ v]$, where $n$ represents the length of the input sequence, $x^t$ represents a text token $x^v$ represents a visual token. To establish the foundational context for efficient decoding, the model computes Key and Value vectors:
\begin{equation}
    K = XW_k, \quad V = XW_v,
\end{equation}
where $W_k\in \mathbb{R}^{d\times d} $ and $W_v\in \mathbb{R}^{d\times d} $ are the weight matrices of the attention module, respectively. $d$ 
represents the hidden dimension of the model. Subsequently, the Key and Value of all tokens are stored in the KV Cache:
\begin{equation}
    \text{KV Cache} = \{K,V\}, 
\end{equation}

\noindent \textbf{Decoding phase:} During iterative token generation, the model leverages the KV Cache and existing context to predict the next token. At generation step $t$, let the input of the attention module be $x^t$. The model firstly computes its corresponding Query, Key and Value projections:
\begin{equation}
   Q_{t} = x_{t}W_q, \quad K_{t} = x_{t}W_k, \quad V_{t} = x_{t}W_v.
\end{equation}
Then, the KV Cache gets updated as:
\begin{equation}
    K=[K,K_{t}],\quad V=[V,V_{t}].
\end{equation}
This expanded cache enables efficient attention computation for the subsequent token prediction. The $t$-th step attention outputs $O_t$ is computed :
\begin{equation}
    O_t = \text{softmax}\left(\frac{Q_{t} \cdot K^T}{\sqrt{d_k}}\right) V
\end{equation}
While the KV Cache mechanism effectively eliminates redundant key-value recomputation during MLLMs inference, its substantial memory footprint necessitates efficient compression strategies.

\subsection{Observation}
In this section, we explore the attention patterns of MLLMs in multimodal long-context scenarios, presenting experimental findings. The study is conducted on the LLaVA-v1.5-7B \cite{liu2024visual} using the Milebench \cite{song2024milebench} benchmark.

Unlike traditional LLMs long-context, multimodal long contexts involve not only interactions among tokens within the same modality but also more complex interactions between tokens of different modalities.    However, there is a lack of comprehensive analysis of modal behaviors within multimodal long contexts. We investigate the modal behaviors within the MLLMs attention patterns in multiple dimensions, including token level, head level of attention, layer level, and task-specific context.

\paragraph{Token level.}
Prior studies \cite{zhang2024h2o,Xiao2023EfficientSL,chen2024nacl} have demonstrated that the attention mechanisms of LLMs exhibit inherent sparsity, characterized by a small portion of tokens contributing the majority of attention scores.  This sparsity allows for the eviction of unnecessary KV embeddings, thereby reducing the memory demands of KV cache.  Building on these insights, we investigate attention mechanism sparsity in MLLMs.  Specifically, we conduct zero-shot inference on two datasets: Text Needle (TN) and Conversational Embodied Dialogue (ED), analyzing attention score distributions under varying token budgets.  To provide a more granular understanding, we partition tokens into modality-specific subsets and quantify the sparsity for each subset independently, thus extending the analysis beyond global token sparsity. All attention scores are aggregated through layer-wise and head-wise averaging.

The findings are depicted in Figure \ref{fig:findings} (a). We observe that the attention mechanisms of MLLMs exhibit significant sparsity: with merely 20\% of tokens capture nearly 90\% of attention scores, which demonstrates that a compact token subset can effectively approximate full KV embeddings.   Notably, modality-specific sparsity patterns exhibit marked differences.   Specifically, textual tokens display sharp attention concentration, whereas visual tokens exhibit diffuse attention allocation.     This observation aligns with our earlier discussion on information density: visual tokens' lower information density correlates with their flatter attention distributions.     This sparsity heterogeneity reveals the limitations of prior KV cache compression methods for LLMs that treat all tokens equally and underscores the need for modality-specific compression strategies to optimize performance.

\paragraph{Attention head level.}
In Figure \ref{fig:findings} (c), we present the attention score matrices of different attention heads (with additional examples provided in the appendix). It can be observed that, similar to LLM, the attention heads of MLLMs exhibit distinct attention patterns for the same sample.     To investigate the behavior of different modalities across attention heads, we analyze the distribution of attention scores for each modality within each head.     As shown in Figure \ref{fig:findings} (b), there are variations in how attention heads allocate attention scores to different modalities. Typically, an attention head tends to favor one modality by assigning it a higher proportion of attention scores.     We refer to this phenomenon as modality preference. Intuitively, modality preference indicates the modality that an attention head is adept at processing. This insight suggests that we should retain more tokens corresponding to the preferred modality.

\paragraph{Layer level.}
The attention patterns of MLLMs across different layers are similar to those of LLM, as illustrated in Figure \ref{fig:findings} (c). In the initial layers of the model, the attention scores are distributed evenly, while in the subsequent layers, they are concentrated on a few tokens. This observation suggests that we should retain more KV embeddings in the initial layers and fewer in the later layers, which aligns with the claims made by \citet{cai2024pyramidkv}. However, \citet{liu2024scissorhands} argue that important tokens exhibit greater variability in the higher layers, thus requiring a larger cache to reduce cache misses. We also observe a corresponding phenomenon in the attention score matrix of MLLMs, where the distribution of attention scores is more uniform in the final layer compared to the middle layers. These observations lead us to recognize the importance of tokens in both the initial and final layers, necessitating the preservation of more KV embeddings.

\paragraph{Task level.}
As previously discussed in the earlier sections, the significance of modalities varies across different tasks. For instance, as illustrated in Figure \ref{taskllevel}, in the Text Needle task, textual tokens hold greater importance, whereas in Image Retrieval, visual tokens dominate. This disparity necessitates the pursuit of a modality-adaptive compression strategy.

\subsection{MadaKV}
In this section, we present a modality-adaptive KV cache compression strategy, encompassing Modality Preference Adaptation and Hierarchical Compression Compensation mechanisms. This approach dynamically adjusts to the varying significance of modalities, ensuring efficient cache management while maintaining performance.

\paragraph{Modality Preference Adaptation.} The modality preference characterizes the system's inherent bias in allocating attention resources across different modalities. This insight suggests that to retain as much contextual information as possible, we should align with the modality preference and evict modality tokens at different granularities. Formally, we define the preference metric as the sum of the importance of modality tokens:
\begin{equation}
w_v = \sum_{i \in X_v} \psi(i), \quad
w_t = \sum_{i \in X_t} \psi(i)
\end{equation}
where $X_v$ and $X_t$ denote the sets for visual and textual tokens respectively. In long-context scenarios, evaluating token importance using cumulative attention can be biased \cite{chen2024nacl}. Here, we opt for proxy tokens to assess token importance more fairly:
\begin{equation}
\psi(i) = \sum_{j \in \mathcal{P}} \alpha_{j \to i}
\end{equation}
where $\mathcal{P}$ denotes the proxy token set, and $\alpha_{j \to i}$ represents the attention score from the proxy token $j$ to token $i$. We select a few tokens from the end of the prompt as proxy tokens, as they typically represent task-specific questions.

We now introduce how to compute the cache budget for each modality using modality preference information. For the $h$-th attention head in the $l$-th layer, the budget for each modality is determined based on preference metric:
\begin{equation}
\varphi^{l,h}_{v} = \frac{w_v}{w_v + w_t} \varphi^{l}, \quad \varphi^{l,h}_{t} = \frac{w_t}{w_v + w_t} \varphi^{l}
\end{equation}
where the cache budget for the $l$-th layer of the model as $\varphi ^l$. Our method conducts modality-specific KV cache eviction for each attention head, effectively improving cache utilization. This flexibility allows our approach to be seamlessly adapted to a wide range of application scenarios.

\paragraph{Hierarchical Compression
Compensation.}
We present a Hierarchical Compression Compensation (HCC) strategy that adaptively modulates token eviction policies across different layers based on the input instance. This design is motivated by the fundamental observation that different layers inherently possess diverse sparsity characteristics, which significantly impacts the model's overall performance.

Prior work has shown that evicting tokens in shallow layers can lead to cascading errors that propagate and amplify through the network \cite{zhang2024lorc}. As a result, conservative eviction is recommended in these layers. Conversely, \citet{liu2024scissorhands} argue that important tokens exhibit greater variability in higher layers, necessitating larger caches to reduce cache misses.
In this work, we integrate these insights from prior research. Rather than empirically predefining the importance of each layer, we propose an adaptive approach that considers both the complexity of modality information in the current layer and the compression status of preceding layers. This ensures that the changes brought about by eviction remain within acceptable bounds, thereby balancing the trade-off between cache efficiency and information retention. Specifically, we define the sparsity of each modality within the attention heads as follows:
\begin{equation}
\begin{gathered}
k_v^{l,h} = \min \Bigl\{ |\mathcal{C}_v| \Bigm| \sum\nolimits_{i\in\mathcal{C}_v} \psi(i) \geq \theta w_v \Bigr\}, \\
k_t^{l,h} = \min \Bigl\{ |\mathcal{C}_t| \Bigm| \sum\nolimits_{i\in\mathcal{C}_t} \psi(i) \geq \theta w_t \Bigr\},
\end{gathered}
\end{equation}
where $\mathcal{C}_v$ and $\mathcal{C}_t$ represent subsets of visual and text tokens respectively, and $\theta$ indicates the threshold value. The budget compensation for the $l$-th layer is defined as follows:
\begin{equation}
    K^l=\sum_{h=1}^{H}(k^{l,h}_v+k^{l,h}_t-\varphi ^l),
\end{equation}
where $H$ denotes the number of attention heads in the model's attention mechanism and $\varphi ^l$ indicates the allocated cache budget for the $l$-th layer. A positive $K^l$ value indicates that the current layer has exceeded its budget, while a negative value means the current layer has a saved budget. This budget compensation is accumulated across layers and influences the budget allocation for subsequent layers:
\begin{equation}
    \varphi^{l+1}=\varphi^{l}-\frac{K^l}{L-l},
\end{equation}
where $L$ denotes the total number of layers in the model. The inter-layer compression compensation allows the model to adaptively adjust its strategy for the current layer based on the modality sparsity of the current layer and the compression status of previous layers.  Compared to prior heuristic approaches, this approach is more generalizable and effective.

\begin{table*}[ht]
\centering
\small
\setlength{\tabcolsep}{5pt}
\begin{tabular}{c|cccccccccc}
\toprule[1.25pt]
\toprule 
Method & TN & IEdit & MMCoQA & STD & ALFRED & CLEVR-C & DocVQA & ST & OI & Average \\
\midrule
\multicolumn{11}{l}{\textbf{\textit{\small{LLaVA-v1.5-7B}}}} \\
\midrule
Full Cache & 9.68 & 7.98 & 33.50 & 16.32 & 15.18 & 16.62 & 46.00 & 63.50 & 48.50 & 28.59 \\
\midrule
StreamingLLM & 3.12 & 3.59 & 26.00 & 11.77 & 3.73 & 10.44 & 42.50 & 43.00 & 44.00 & 20.91 \\
H2O & 2.50 & 5.51 & 28.00 & 15.73 & 14.86 & 14.07 & 44.00 & 63.50 & 44.00 & 25.80 \\
SnapKV & 3.27 & 6.03 & 29.00 & 14.82 & 14.40 & 15.37 & 45.50 & \textbf{64.00} & 45.50 & 26.43 \\
LOOK-M & 3.34 & 6.51 & 29.50 & 15.79 & 13.96 & 14.12 & 45.50 & 63.00 & 46.50 & 26.47 \\
MadaKV & \textbf{9.38} & \textbf{6.97} & \textbf{31.00} & \textbf{15.85} &\textbf{15.06} & \textbf{16.76} & \textbf{47.00} & \textbf{64.00} & \textbf{48.00} & \textbf{28.22} \\
\midrule
\multicolumn{11}{l}{\textbf{\textit{\small{LLaVA-v1.5-13B}}}} \\
\midrule
Full Cache & 25.34 & 9.01 & 40.50 & 15.70 & 18.56 & 15.98 & 55.50 & 74.50 & 52.00 & 34.12 \\
\midrule
StreamingLLM & 7.81 & 2.63 & 28.50 & 14.05 & 1.72 & 6.55 & 49.80 & 68.00 & 50.00 & 25.45 \\
H2O & 7.81 & 8.70 & 34.50 & 15.21 & 15.03 & 15.06 & 53.00 & 74.00 & 51.00 & 30.48 \\
SnapKV & 7.92 & 8.52 & 35.00 & 15.53 & 16.84 & 15.37 & 53.50 & \textbf{74.50} & 51.00 & 30.91 \\
LOOK-M & 10.00 & 8.77 & 37.50 & 15.46 & 15.25 & 14.12 & 53.00 & 74.00 & 50.50 & 30.96 \\
MadaKV & \textbf{23.43} & \textbf{9.41} & \textbf{38.50} & \textbf{15.62} & \textbf{17.76} & \textbf{16.76} & \textbf{55.00} & \textbf{74.50} & \textbf{51.50} & \textbf{33.61} \\
\midrule
\multicolumn{11}{l}{\textbf{\textit{\small{Qwen2.5-VL-7B}}}} \\
\midrule
Full Cache & 11.25 & 29.45 & 44.50 & 28.36 & 37.53 & 42.46 & 62.50 & 63.00 & 61.00 & 63.34 \\
\midrule
StreamingLLM & 3.46 & 25.12 & 40.00 & 26.14 & 34.07 & 27.58 & 60.00 & 59.00 & 60.50 & 55.98 \\
H2O         & 3.97 & 29.13 & 40.50 & 26.79 & 37.78 & 36.03 & \textbf{62.50} & 61.50 & 61.00 & 59.70 \\
SnapKV      & 4.55 & 29.25 & 41.50 & 26.54 & 37.41 & 39.41 & 61.50 & 60.00 & 60.50 & 60.11 \\
LOOK-M      & 4.23 & 28.73 & 42.00 & \textbf{28.97} & 36.66 & 38.66 & 62.00 & 61.50 & 61.50 & 60.71 \\
MadaKV      & \textbf{10.73} & \textbf{29.39} & \textbf{43.50} & 27.30 & \textbf{37.95} & \textbf{40.57} & \textbf{62.50} & \textbf{62.00} & \textbf{62.50} & \textbf{62.74} \\
\bottomrule[1.25pt] 
\end{tabular}
\caption{Performance of eviction strategies.  The best results are highlighted in \textbf{bold}.}
\label{main_result}
\vspace{-0.5cm}
\end{table*}

\section{Experiments}

\subsection{Setting}
We sample nine tasks from the MileBench benchmark \cite{song2024milebench}, which is the first benchmark specifically designed to test the long-context multimodal capabilities of MLLMs.   MileBench covers a wide range of general scenarios, including temporal multi-image tasks, semantic multi-image tasks, needle-in-a-haystack tasks, and image retrieval tasks.   On average, each sample in MileBench contains 15.2 images and 422.3 words.

To evaluate MadaKV, we conduct comprehensive experiments on widely-adopted long-context MLLMs: LLaVA-v1.5-7/13B \cite{liu2024visual} and Qwen2.5-VL-7B. 
We compare MadaKV against several representative KV cache eviction methods.    We include StreamingLLM \cite{xiao2023efficient}, H2O \cite{zhang2024h2o}, and SnapKV \cite{chen2024nacl}, which are all text-based KV cache eviction methods.      Additionally, we compare against LOOK-M \cite{wan2024look}, an eviction method designed for multimodal scenarios.

\subsection{Experiment Results}

In Table \ref{main_result}, we present a comprehensive comparison of MadaKV against various eviction methods in the context of multimodal long-context scenarios. The experimental results highlight MadaKV's effectiveness in managing KV caches under memory constraints while maintaining high performance across diverse tasks. Specifically, MadaKV achieves an 80\% reduction in memory usage, with only a slight drop in average accuracy compared to full caching. This demonstrates MadaKV's ability to significantly reduce memory footprint with minimal performance trade-offs.

Compared to baseline eviction methods, MadaKV consistently outperforms across most datasets.   Notably, MadaKV substantially outperforms single-modality text-based KV eviction methods.   For instance, in the TN task, MadaKV achieves a 6.11\% improvement in performance.   This finding underscores a critical limitation of text-based KV eviction methods in multimodal long-context scenarios: they often overlook modality-specific information, leading to improper eviction of KV caches and generating erroneous responses.
Moreover, MadaKV outperforms LOOK-M, a method designed for multimodal scenarios.   Unlike LOOK-M, which prioritizes retaining text tokens, MadaKV adaptively determines the importance of tokens from different modalities based on the specific task context.   This approach prevents excessive eviction of tokens from any single modality, thereby enhancing the model's robustness when handling complex multimodal long-context inputs.

\begin{figure*}[th]
  \centering
  \includegraphics[width=16.01cm]{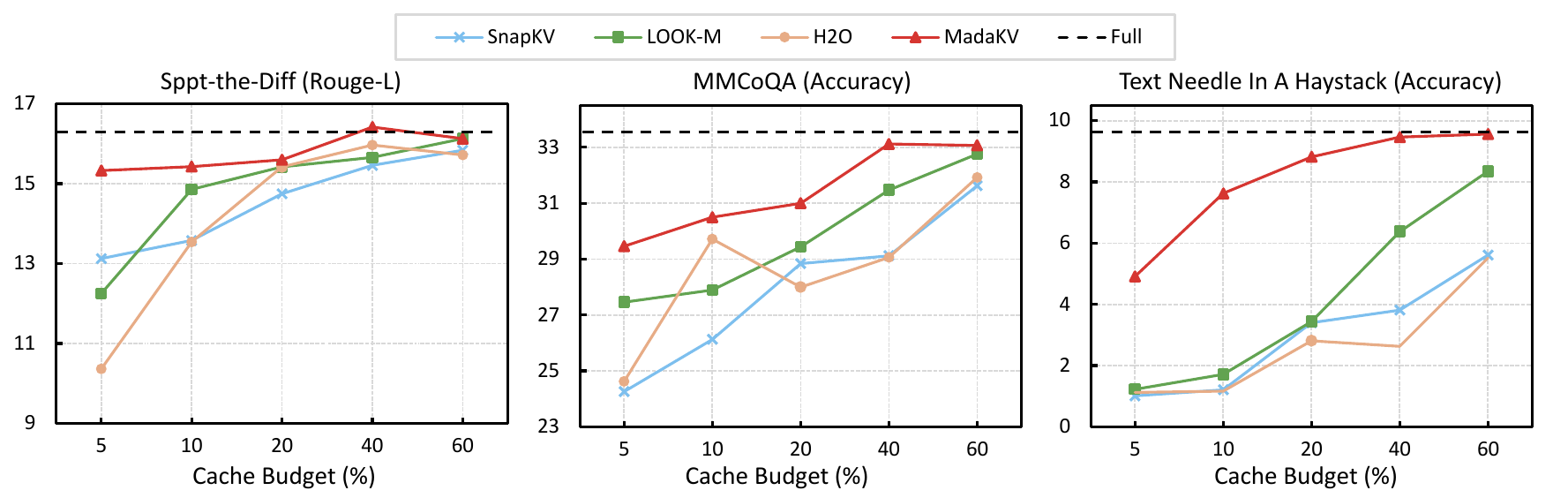}
  \caption{Comparison results for various cache budgets.}
  \label{zhexian}
\end{figure*}

\subsection{Influence of Various Cache Budgets }
To evaluate the effectiveness of MadaKV under varying cache budgets, we conduct experiments on the LLava-v1.5-7B model with cache budgets ranging from 5\% to 60\%.    We select three subtasks for assessment: Spot-the-Diff, MMCoQA, and Text Needle In A Haystack.    The results are presented in Figure \ref{zhexian}. Most achieve methods performance comparable to full caching when using a 50\% cache budget.    This suggests that multimodal long-context scenarios contain substantial redundant information, which can be pruned to reduce memory usage without significant performance loss.  MadaKV consistently outperforms the baselines across all cache budgets.    Notably, in the Text Needle In A Haystack task, MadaKV's performance with a 20\% cache budget matches that of LOOK-M using a 60\% cache budget.    Additionally, MadaKV shows significant improvements over baselines when the cache budget is below 10\%.    These findings demonstrate MadaKV's ability to accurately identify critical information within KV caches, thereby minimizing context loss even under stringent memory constraints.

\subsection{Efficiency Analysis}

\begin{table}[t]
\centering
\scalebox{0.8}{
\begin{tabular}{lccc}
\toprule[1.25pt]
\toprule 
Method & Budget & Decoding Latency & GPU Memory \\ 
\midrule
Full Cache & 100\% &  27.85 ms/token & 1.63 GiB \\ 
MadaKV  & 20\% &  19.57 ms/token & 0.41 GiB \\
MadaKV  & 5\% & \textbf{17.16 ms/token} & \textbf{0.16 GiB} \\
\bottomrule[1.25pt]
\end{tabular}
}
\caption{Model Speed and KV Cache GPU Memory Usage. The best results  are highlighted in \textbf{bold}.}
\label{tab:model_performance}
\vspace{-0.5cm}
\end{table}

We delve into the efficiency of our proposed method, as detailed in Table \ref{tab:model_performance}. Specifically, we examine the decoding speed and memory usage of model inference both with and without our method. To ensure the reliability and robustness of our findings, we conduct tests on decoding latency and GPU memory usage using 20 randomly selected data entries. All speed tests were performed on a single NVIDIA A100 Tensor Core GPU.

As shown in Table \ref{tab:model_performance}, MadaKV exhibits significantly lower decoding latency compared to the model with a full cache. This advantage is particularly evident in long generation tasks, where the efficiency of our method is further accentuated.
 Moreover, we analyze the speed and GPU memory usage of KV Cache under two budget scenarios: 20\% and 5\%. These results were derived from the mean values obtained during the inference process of 20 randomly sampled data points. Our findings reveal that the average GPU memory consumption is nearly proportional to the cache budget. Specifically, at a 20\% KV Cache budget, memory usage during the decoding stage is reduced by approximately 80\% compared to a Full Cache scenario. This highlights the substantial memory savings achieved through our method.

\begin{table}[t]
\setlength{\tabcolsep}{9pt}
    \centering
    \begin{tabular}{cc|ccc}
    \toprule[1.25pt]
    \toprule 
        MPA & HCC & TN & IEdit & ALFRED  \\ 
        \midrule
        \ding{56} & \ding{56} & 2.47 & 3.55 & 14.32 \\ 
        \ding{52} & \ding{56} & 6.58 & 5.72 & 14.86 \\ 
        \ding{56} & \ding{52} & 5.51 & 5.19 & 14.61 \\ 
        \ding{52} & \ding{52} & \textbf{9.38} & \textbf{6.97} & \textbf{15.06} \\ 
        \bottomrule[1.25pt]
    \end{tabular}
    \caption{Ablation study of the effect of individual module. MPA: modality preference adaptation, HCC: hierarchical compression compensation. The best results  are highlighted in \textbf{bold}.}
    \label{tab:ablation}
    \vspace{-0.5cm}
\end{table}

\subsection{Ablation Study}
\paragraph{The Effect of Modality Preference Adaptation.} Modality Preference Adaptation is used to determine the modality preferences of attention heads, indicating which modality each head tends to allocate more attention scores to.    This information guides the model in deciding how many tokens to retain for each modality.    In Table \ref{tab:ablation}, we report the impact of MPA on model performance.    Removing this strategy leads to a noticeable drop in performance.
These results demonstrate the effectiveness of MPA in multimodal long-context scenarios.    As previously discussed, the importance of different modalities varies greatly across tasks, making it impractical to simply designate one modality as universally more critical than others.    MPA allows the model to dynamically adjust the number of tokens retained for each modality based on its actual importance.    This dynamic adjustment improves the model's ability to integrate multimodal information while maintaining computational efficiency.

\paragraph{The Effect of Hierarchical Compression Compensation.} As illustrated in Table \ref{tab:ablation}, Hierarchical Compression Compensation significantly enhances model performance.  For instance, in the TN task, HCC leads to a 3.07\% improvement, underscoring its effectiveness.  This finding highlights the importance of allocating different cache budgets to different layers in multimodal long-context scenarios.
Our approach dynamically adjusts cache resource allocation by considering both the sparsity of the current layer and the cache budget usage of previous layers.  This strategy not only effectively reduces redundant computation but also ensures that critical information is retained and efficiently processed throughout the long context.

\section{Conclusion}
In this study, we introduce MadaKV, a modality-adaptive KV cache eviction strategy designed to optimize the inference efficiency of MLLMs in long-context scenarios. MadaKV leverages modality preferences to guide the granularity of token eviction by attention heads and employs inter-layer compensation to dynamically adjust eviction ratios across layers based on overall modality complexity. Through extensive experiments on a diverse set of multimodal long-context tasks using the representative MileBench benchmark, we demonstrate the effectiveness of MadaKV.
Looking ahead, we plan to explore the integration of MadaKV with other MLLMs inference acceleration techniques to further enhance efficiency and performance.

\section{Limitation}
The primary limitations of our approach are as follows: First, due to resource constraints, we have not yet conducted experiments on models with larger parameter sizes (e.g., 34B, 70B) or on datasets with extremely long contexts. However, based on our current experimental analysis, MadaKV appears to be scalable and adaptable to a variety of application scenarios. Additionally, this study has focused on the common visual and textual modalities within MLLMs and has not yet explored other modalities (e.g., video, audio). These areas will be prioritized as future directions for exploration.

\section{Acknowledgements}
This work was supported by the Key Research and Development Program of Zhejiang Province (No. 2024C03270), and the National Natural Science Foundation of China (No. 62402429, U24A20326, 62441236). This work was also partially supported by the ZJU Kunpeng\&Ascend Center of Excellence, the Ningbo Yongjiang Talent Introduction Programme (No. 2023A-397-G) and the Young Elite Scientists Sponsorship Program by CAST (No. 2024QNRC001).

\bibliography{custom}



\appendix

\section{Appendix}
\label{sec:appendix}
\begin{figure}[h]
  \centering
  \includegraphics[width=7.51cm]{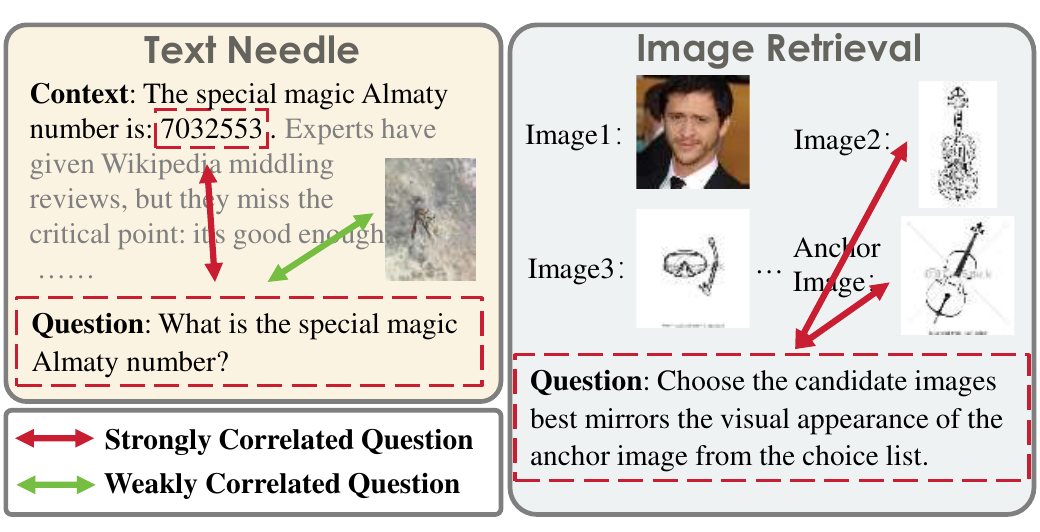}
  \caption{Examples of modal significance in different tasks.}
  \label{taskllevel}
\end{figure}
\subsection{The Significance of Modalities Across Different Tasks}
We provide an example to illustrate the substantial differences in modality importance across various tasks. As shown in Figure \ref{taskllevel}, the Text Needle task focuses on locating the corresponding answer within the text, while the Image Retrieval task centers on comparing the similarity between images. These examples highlight how the significance of modalities varies significantly depending on the specific task requirements.

\subsection{Details of MileBench}
MileBench \cite{song2024milebench} dataset is the first benchmark specifically designed to test the Multimodal Long-context capabilities of MLLMs. Milebench primarily includes 6,440 multimodal long-text samples, which are composed of 21 existing or self-constructed datasets, with an average of
15.2 images and 422.3 words per sample.
The detailed information of dataset is presented in Table \ref{tab:dataset}.

\begin{table*}[t]
\small
\setlength{\tabcolsep}{9pt}
\renewcommand{\arraystretch}{1.1}
    \centering
    \begin{tabular}{c|c|c}
    \toprule[1.25pt]
    \toprule 
        \textbf{Dataset Abbr.} & \textbf{Task} & \textbf{Data Source} \\ 
        \midrule
        TN & Text Needle In A Haystack& TextNeedleInAHaystack \\ 
        IEdit & Visual Relationship Expressing & IEdit~\citep{IEdit}\\ 
        MMCoQA & Multimodal Dialogue & MMCoQA~\citep{MMCoQA} \\ 
        STD & Visual Change Captioning & Spot-the-Diff~\citep{Spot-the-Diff} \\ 
        ALFRED & Conversational Embodied Dialogue &  ALFRED~\citep{ALFRED} \\
        CLEVR-C & Visual Change Captioning & CLEVR-Change~\citep{CLEVR-Change} \\
        DocVQA & Document QA & DocVQA~\citep{DocVQA}  \\
        ST & Scene Transition & MovieNet~\citep{MovieNet} \\
        OI & Object Interaction & STAR~\citep{STAR} \\
        \bottomrule[1.25pt]
    \end{tabular}
    \caption{Detailed Statistics and Taxonomy of dataset.}
    \label{tab:dataset}
\end{table*}

\begin{figure*}[h]
  \centering
  \includegraphics[width=16.01cm]{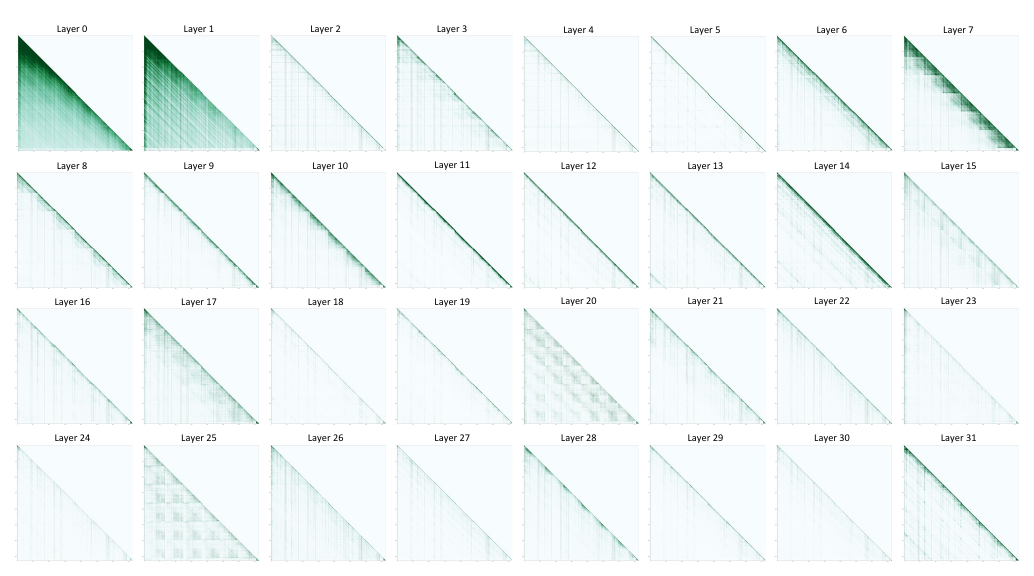}
  \caption{Attention patterns of LLava-v1.5-7B on ALFRED.}
  \label{al_attenti}
\end{figure*}

\end{document}